\documentclass[conference]{IEEEtran}
\IEEEoverridecommandlockouts
\usepackage{subcaption}
\usepackage{caption}
\usepackage{amsmath,epsfig}
\usepackage{bm}
\usepackage{slashbox}
\usepackage{algorithmic}
\usepackage{latexsym}
\usepackage{amsmath}
\usepackage{amssymb}
\usepackage{epsfig}
\usepackage{amsthm}
\usepackage{graphicx}
\usepackage[ruled,vlined]{algorithm2e}
\usepackage{epstopdf}

\begin{document}
\title{{ \huge Non-parametric Bayesian Learning with Deep Learning Structure and Its Applications in Wireless Networks}}

\author{Erte Pan and Zhu Han \thanks{E. Pan (epan@uh.edu), and Z. Han (zhan2@uh.edu) are with the Department of Electrical and Computer Engineering, University of Houston, Houston, Texas, USA, 77004.}}


\maketitle\thispagestyle{empty}

\begin{abstract}
In this paper, we present an infinite hierarchical non-parametric Bayesian model to extract the hidden factors over observed data, where the number of hidden factors for each layer is unknown and can be potentially infinite. Moreover, the number of layers can also be infinite. Previous non-parametric Bayesian methods assume binary values for both hidden factors and weights. In contrast, we construct the model structure that allows continuous values for the hidden factors and weights, which makes the model suitable for more applications. We use the Metropolis-Hastings method to infer the model structure. Then the performance of the algorithm is evaluated by the experiments. Simulation results show that the model fits the underlying structure of simulated data.
\end{abstract}

\begin{IEEEkeywords}
non-parametric Bayesian learning, deep learning, Indian Buffet Process, Metropolis-Hastings algorithm
\end{IEEEkeywords}

\section{Introduction}
\label{sec:IntroBayesianHidden}
Statistical models have been applied to the classification and prediction problems in machine learning and data analysis \cite{TehJor2010a}. Some statistical methods make the hypothesis of mathematical models that are controlled by certain parameters to fit the latent structure of observed data \cite{Thibaux07hierarchicalbeta}. The observed data are assumed to be generated by complex structures which have hierarchical layers and hidden causes \cite{Rasmussen00theinfinite}. One key challenge faced by modeling the data structure in this way is thus the determination of the numbers of layers and hidden variables. However, it is sometimes impractical and challenging to choose any fixed number for the model structure when making the hypothesis. Therefore, we need flexible non-parametric models that make fewer assumptions and are capable of an unlimited amount of latent structures. Hierarchical nonparametric Bayesian model assumes unspecified number of latent variables and produces rich kinds of probabilistic structures by constructing cascading layers. Hence, it is considered to be a powerful technique to cope with the challenge.

In \cite{Wood06anon-parametric}, a two-layer non-parametric Bayesian model was proposed with both hidden factors and linking weights being binary. The model accommodates potentially infinite number of hidden factors and performs well in inferring stroke localizations. Works in \cite{AdamsGraphical10} built a deep cascading graphical model that permits the number of hidden layers to be infinite. This technique has been used in the inference of the structures of images. However, the proposed model only infers the priors of the number of hidden factors in each layer and ignores the influence of factor values in each layer on the posterior distributions of factor numbers. In \cite{ICML2011Chen251}, the authors developed a hierarchical model based on the Beta process for convolutional factor analysis and deep learning. In the proposed linear model, the connecting weights and hidden factors are real values other than binary ones. The model has been used in multi-level analysis of image-processing data sets. In \cite{rai08ihfrm}, another approach was constructed to build the prior distribution for the nonparametric Bayesian factor regression. The Kingman's Coalescent is chosen as the prior and achieves good results in gene-expression data analysis. In \cite{Knowles07}, the Indian Buffet Process (IBP) was introduced into factor analysis and therefore enabled their model of handling the infinite case. In addition, the method allows real-valued weights and factors. In the application realm, Non-parametric Bayesian model has been explored to solve various classification and clustering problems. In \cite{Lee09convDBNAudio}, the deep belief networks had been applied to unlabeled auditory data and achieved good performance in the unsupervised classification task. Although the nonparametric Bayesian technique has been advanced by researchers recently, challenges still lie in the problem of constructing the real-valued non-linear models with the numbers of both hidden layers and hidden factors being infinite.

In this paper, we investigate the nonparametric Bayesian graphical model with infinite hierarchical hidden layers and an infinite number of hidden factors in each layer. Our main contributions include: the proposition of the infinity structure both latently and hierarchically; the linking weights are extended from binary values to real values; the proposed model is constructed in a non-linear fashion, not like the works in \cite{ICML2011Chen251}; the employment of the Metropolis-Hastings algorithm enables the alternative update of the values of hidden factors layer by layer, making the inference procedure recursively. The phantom data are simulated according to our infinite generative model. The inferring algorithm is then applied to the simulated data to extract the data structure. As is stated before, when considering wireless security circumstance, the applications we mainly focus on turn to be the clustering problems. Therefore, the most interest lies in the number of hidden factors which indicates the number of clusters in different hierarchical levels. The simulation results show that this greedy algorithm accomplishes the objective of discovering the number of hidden factors accurately.

This paper is organized as follows: In Section \ref{sec:GenerativeModelBayesianHidden}, the nonparametric Bayesian generative model is introduced to generate the data. The inference algorithm is given in Section \ref{sec:InferenceAlgorithmBayesianHidden}. Simulation results are presented in Section \ref{sec:SimulationBayesianHidden}. In Section \ref{sec:ConclusionsBayesianHidden}, we draw conclusions and give insightful discussions.

\section{Generative Model}
\label{sec:GenerativeModelBayesianHidden}
The objective is to construct a hierarchical Bayesian framework based generative model which allows both infinite layers and infinite components in each layer. To better explain the proposed model, we describe the finite generative model first. Then the infinite model can be obtained by extending the number of hidden factors and the number of layers to infinity.

\subsection{Finite Generative Model}
Finite generative model is used to model the causal effects among the factors between layers, as those described in \cite{Wood06anon-parametric} and \cite{NamCellularBook}. Here we construct the model of one observation layer and two hidden layers. Define the matrix $ \mathbf{X} = [\mathbf{x}_{1},\mathbf{x}_{2},...,\mathbf{x}_{T}]$ as the data set of $T$ data points with each $\mathbf{x}_{t}$ being a vector of $N$ dimensions. Accordingly, define the matrix $\mathbf{Y}^1 = [\mathbf{y}^1_{1},\mathbf{y}^1_{2},...,\mathbf{y}^1_{T}]$ as the hidden factors of first hidden layer with each $\mathbf{y}^1_{t}$ being a vector of $K^1$ dimensions. Similarly, we have the definition for the $K^2 \times T$ matrix $\mathbf{Y}^2$ as the hidden factors of second hidden layer. To express the dependency between two successive layers, we use the $N \times K^1$ weight matrix $\mathbf{W}^{1}$ and $K^1 \times K^2$ weight matrix $\mathbf{W}^{2}$, respectively. For instance, if there exists a connection between $\mathbf{Y}^1_{k,t}$ and $\mathbf{X}_{n,t}$, which means the hidden cause $\mathbf{Y}^1_{k,t}$ will influence the generation of data component $\mathbf{X}_{n,t}$, then $\mathbf{W}^{1}_{n,k}\neq{0}$ and $\mathbf{W}^{1}_{n,k}\in{\Re}$. Otherwise, $\mathbf{W}^{1}_{n,k}=0$. The rest hidden vectors $\{\mathbf{y}^i\}$ and weight matrices $\{\mathbf{W}^i\}$ can be derived in the similar way. Fig. 1 illustrates the proposed infinite generative model structure for a particular instance $t\in{\{1,2,...,T\}}$. Note that the weight matrices remain the same through all instances ${\{1,2,...,T\}}$ while the data sets between two instances are generated independently.

\begin{figure}
\label{fig:modelBayesianHidden}
\centering
\includegraphics[width=75mm]{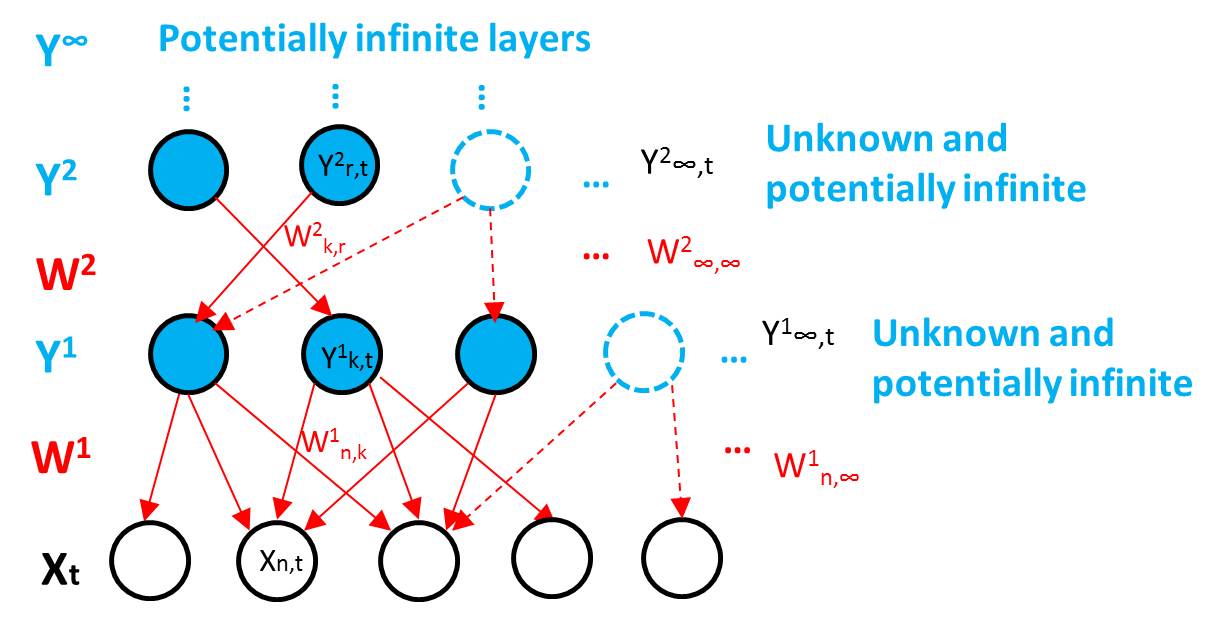}
\caption{Proposed infinite generative model}
\label{fig:infite model}
\end{figure}

Within one particular instance $t$, data vector $\mathbf{x}_{t}$ is generated in as follows: First, the hidden vector $\mathbf{y}^i_{t}$ of the topmost layer is generated according to Gaussian distribution $N(0,\sigma^{2}_{y,i})$. Then the weight matrix $\mathbf{W}^{i}$ is generated according to $\mathbf{W}^{i}=\mathbf{Z}^{i}\bigodot{\mathbf{G}^{i}}$, where matrices $\mathbf{Z}^{i}$ and $\mathbf{G}^{i}$ are of the same size as $\mathbf{W}^{i}$ and the symbol $\bigodot$ indicates the element-wise product operator. We assume each column of matrix $\mathbf{Z}^{i}$, $\mathbf{Z}^{i}_{.,r}$, is generated independently as $\mathbf{Z}^{i}_{.,r}\sim{Bernoulli(p_{r})}$. We further impose a prior distribution for the parameter $p_{r}\sim{Beta(\alpha'_{i}/K^i,1)}$. It will be demonstrated later that this strategy of constructing matrix $Z^{i}$ will result in the Indian Buffet Process as the number of variables of $\mathbf{y}^i_{t}$, $K^i$, approaches the infinity \cite{Griffiths11IBP}. For the matrix $\mathbf{G}^{i}$, each column is generated by Gaussian distribution $\mathbf{G}^{i}_{.,r}\sim{N(0,\sigma^{2}_{r})}$ with the variance conforms to the inverse gamma prior, $\sigma^{2}_{r}\sim{InverseGamma(\alpha_{2},\beta_{2})}$. The matrix $\mathbf{Z}^{i}$ imposes the selection effect of variables between layers while the matrix $\mathbf{G}^{2}$ indicates how much influence a variable will receive from its higher level variables or ancestors. Having obtained the hidden vector $\mathbf{y}^i_{t}$ and weight matrix $\mathbf{W}^{i}$, the variables of $\mathbf{y}^{i-1}_{t}$ are conditionally independently generated given $\mathbf{y}^i_{t}$ and $\mathbf{W}^{i}$, and we assume they follow the Gaussian distribution $\mathbf{Y}^{i-1}_{k,t} \sim N(0,\sigma^{2}_{y,i-1})$, where the parameter $\sigma_{y,i-1}$ is specified by $\sigma_{y,i-1}=\left|{\sum_{j=1}^{K^i} W_{k,j}^{i} Y^i_{j,t}}\right|$. It can be verified that the element of weight matrix $\mathbf{W}^{i}$ follows the distribution:
\begin{equation}
\begin{split}
P(W_{k,r}^i | p_{r}, \sigma^{2}_{r} ) & = \left|{sgn(W_{k,r}^i)}\right|p_{r}N(W_{k,r}^i;0,\sigma^{2}_{r})\\
& +(1-p_{r})\delta_0(W_{k,r}^i),
\end{split}
\end{equation}
where $sgn$ indicates the sign function and symbol $\delta_0$ is a delta function at 0. The downward layers are constructed in the same fashion.

The generation of variables from the first hidden layer $\mathbf{y}^1_{t}$ to the observed layer $\mathbf{x}_{t}$ is similar to the procedure above, except for the parameterizations: We assume $\mathbf{Z}^{1}_{.,k}\sim{Bernoulli(p_{k})}$ and $p_{k}\sim{Beta(\alpha'_{1}/K,1)}$ for the matrix $\mathbf{Z}^{1}$. For the matrix $\mathbf{G}^{1}$, $\mathbf{G}^{1}_{.,k}\sim{N(0,\sigma^2_{k})}$ and $\sigma_{k}\sim{InverseGamma(\alpha_{1},\beta_{1})}$. Hence, the distribution of observed data vector can be expressed as $\mathbf{X}_{n,t} \sim N(0,\sigma^{2}_{xn})$ and $\sigma_{xn}=\left|{\sum_{j=1}^{K} \mathbf{W}_{n,j}^{1} \mathbf{Y}^1_{j,t}}\right|$, where
\begin{equation}
\begin{split}
P(\mathbf{W}_{n,k}^1 | p_{k}, \sigma^{2}_{k} ) & = \left|{sgn(\mathbf{W}_{n,k}^1)}\right|p_{k}N(\mathbf{W}_{n,k}^1;0,\sigma^{2}_{k})\\
& +(1-p_{k})\delta_0(\mathbf{W}_{n,k}^1).
\end{split}
\end{equation}

This generative model can be employed in many applications since we are able to extract not only the features from data points but also the higher level hyper-features from the extracted features. Instances can be found in applications such as human face recognition where the input data are images of human faces with first level of features being curves and edges, second level of features being organs like eyes and nose \cite{FaceBayesianCVPR2012}. Moreover, we allow one variable to possess more than one hidden causes (not like the Infinite Gaussian Mixture Model) which makes the model more robust. In addition, we assume real weight matrix instead of binary ones, and this will bring our model closer to practice since different hidden causes are reasonably weighted.

\subsection{Infinite Generative Model}
Having established the finite generative model, the case of an infinite number of layers can be expressed by the recursive equations:
\begin{equation}
\small \sigma_{y^{i} j}=\left|{\sum_{l} \mathbf{W}^{i+1}_{j,l} \mathbf{Y}^{i+1}_{l,t}}\right|,
\end{equation}
\begin{equation}
\mathbf{Y}^{i}_{j,t} \sim N(0,\sigma_{y^{i} j}^2).
\end{equation}

Moreover, the infinite number of components in each layer can be obtained by taking the limit as $K^i \rightarrow \infty $. We will demonstrate that the distributions on the selection matrices correspond to the IBP. For example, for our assumptions on $\mathbf{Z}^1$, we have:
\begin{equation}
\small P(\mathbf{Z}^1 | \textbf {p})= \prod_{k=1}^K \prod_{n=1}^N P(\mathbf{Z}^1_{n,k}|p_k) =  \prod_{k=1}^K p_k^{m_k} (1-p_k)^{N-m_k},
\end{equation}
where $m_k = \sum_{n=1}^N Z^1_{n,k}$ is the number of data points that select hidden factor $\mathbf{Y}^1_{k,.}$ and $\mathbf{Z}^1$ is a $N \times K$ matrix. Since we place a prior distribution $Beta(\alpha'_{1}/K^1,1)$ on $p_k$, we can integrate out the parameter $\textbf {p}$ to obtain:
\begin{equation}
P(\mathbf{Z}^1)= \prod_{k=1}^K \frac{\frac{\alpha'_{1}}{K} \Gamma(m_k+\frac{\alpha'_{1}}{K}) \Gamma(N-m_k+1)}{ \Gamma(N+1+\frac{\alpha'_{1}}{K}) }.
\end{equation}
By defining the equivalent-class of matrix $\mathbf{Z}^1$ \cite{Griffiths11IBP}, we can find the distribution on $\mathbf{Z}^1$ as $K \rightarrow \infty $:
\begin{equation}
P(\mathbf{Z}^1)= \frac {{\alpha'}_1^{K_h}}{\displaystyle \prod_{n=1}^N K_1^n!} e^{-{\alpha'}_1 H_N} \prod_{k=1}^{K_h}\frac{(N-m_k)!(m_k-1)!}{N!},
\end{equation}
where $K_1^n$ is the number of first hidden layer factors being selected by the $n$-th variable of the data point $X_{.,t}$, $H_N$ is the harmonic number with $H_N=\sum_{j=1}^N \frac{1}{j}$ and $K_h$ is the number of first hidden layer factors selecting $h$ components of the data point. This distribution corresponds to a stochastic process, the IBP \cite{Griffiths11IBP}, which is the analog of dishes selecting by $N$ customers at an Indian Buffet restaurant. The restaurant provides customers (variables of a data point) an infinite array of dishes which corresponds to the infinite components of first hidden layer factors. The first customer tries $Poisson(\alpha'_{1})$ dishes. The succeeding customers select dishes one by one in the way that they firstly select previously selected dishes with probability $m_{-i,k}/i$, where $m_{-i,k}$ is the number of customers who have chosen the $k$-th dish except the $i$-th customer himself. The $i$-th customer then selects next $Poisson(\alpha'_{1}/i)$ new dishes.

\section{Inference Algorithm}
\label{sec:InferenceAlgorithmBayesianHidden}

Having constructed the infinite generative model, the goal is to infer the number of hidden layers as well as the number of hidden factors in each hidden layer based on Bayesian inference. The task is done once we obtain the inference of $\{\mathbf{W}^1,\mathbf{W}^2,...,\mathbf{Y}^1,\mathbf{Y}^2,...\}$ given observed data $\mathbf{X}$. However, direct estimation of $P(\mathbf{W}^1,\mathbf{W}^2,...,\mathbf{Y}^1,\mathbf{Y}^2,...|\mathbf{X})$ is intractable. Inspired by \cite{Hinton06FLA}, we perform the inference one layer at a time. That is, we first initialize the weights matrices $\{\mathbf{W}^1,\mathbf{W}^2,...\}$ as well as the hidden layer $\{\mathbf{Y}^1,\mathbf{Y}^2,...\}$. Then we fix the value of $\{\mathbf{W}^2,...\}$ and $\{\mathbf{Y}^2,...\}$, leading to the fact that the prior distribution of $\mathbf{Y}^1$ is known and can be expressed in terms of $\{\mathbf{W}^2,...\}$ and $\{\mathbf{Y}^2,...\}$. Based on this scenario, we use the Metropolis-Hastings algorithm as an approximate method to infer the first hidden layer $\{\mathbf{Y}^1,\mathbf{W}^1\}$. After inferring first hidden layer, we use matrix $\mathbf{Y}^1$ as the input data points and perform Bayesian inference at the second hidden layer, and so forth. Since the prior has been changed during the inference of second hidden layer, we need to re-infer the first hidden layer using the updated upper hidden layers. We iteratively perform inference at each layer until the value of $\{\mathbf{W}^1,\mathbf{W}^2,...,\mathbf{Y}^1,\mathbf{Y}^2,...\}$ converges. It has been proved that this layer-wise inferring strategy is efficient in \cite{carreiraperpin:aistats2005}. Different from \cite{carreiraperpin:aistats2005}, the Metropolis-Hastings algorithm is applied to perform the inference, instead of the contrastive divergence method.

The Metropolis-Hastings algorithm was first introduced by the classic paper by Metropolis, Rosenbluth etc. in 1953 and has been extensively applied in statistical problems. It defines a Markov chain which allows the change of dimensionality between different states of the model. The new state is generated from the previous state by first generating a candidate state using a specified proposal distribution. Then a decision is made to accept the candidate state or not, based on its probability density relative to that of the previous state, with respect to the desired invariant distribution, $Q$. If the candidate state is adopted, it evolves as the next state of the Markov chain; otherwise, the state of the model stays the same. To better explain the inference algorithm, we specify the problem into one hidden layer inference. The generalized infinite case can be derived in the similar fashion. In our problem settings, let $\eta$ represent the values of $\mathbf{W}^1,\mathbf{Y}^1,K^1$, where $\mathbf{W}^1$ is the weight matrix connecting the $N \times T$ data matrix $\mathbf{X}$ and the $K^1 \times T$ hidden factors matrix $\mathbf{Y}^1$ and $K^1$ is the dimension of hidden factor $\mathbf{y}^1_t$. Then the change between different states of the model is adopted with probability
\begin{equation}
A(\eta^*,\eta) = min\left[1, \frac{P(\mathbf{X},\eta^*)}{P(\mathbf{X},\eta)} \frac{Q(\eta|\eta^*)}{Q(\eta^*|\eta)}\right],
\end{equation}
where $\eta^*$ is the proposed new value, $\eta$ is the current value, and $Q(\eta^*|\eta)$ is the probability of proposing $\eta^*$ given $\eta$. The term $P(\mathbf{X},\eta)$ can be further expressed as
\begin{equation}
P(\mathbf{X},\eta) = P(\mathbf{X}|\mathbf{W}^1,\mathbf{Y}^1) P(\mathbf{Y}^1|K^1) P(\mathbf{W}^1|K^1) P(K^1).
\end{equation}
The change of dimensionality is completed in this way: Iteratively pick a hidden factor with corresponding column $k$ of $\mathbf{W}^1$ and check the number of linked edges $m_k$. If $m_k = 0$, then remove this hidden factor together with the corresponding column in $\mathbf{W}^1$ and decrease $K^1$. Otherwise, propose a new hidden factor with no linked edges and sample the new values of $\mathbf{Y}^1$ by (4).

This new proposed state is accepted with the probability $A(\eta^*,\eta)$. The probability of adding a new hidden factor is approximated by $K^1_+/K^1$ while the probability of generating the new $\mathbf{Y}^1$ is specified by its normal distribution. $Q(\eta^*|\eta)$ is obtained by multiply these two probabilities. To return to the previous configuration, we can delete any hidden factor with the same values as the proposed new row of $\mathbf{Y}^1$. The probability of choosing such a hidden factor is approximated by $1/(K^1+1)$. Therefore, we have
\begin{equation}
\frac{Q(\eta|\eta^*)}{Q(\eta^*|\eta)} = \frac{ 1/(K^1+1)  }{ \frac{K^1_+}{K^1} \prod_{t} \frac{1}{\sqrt{2\pi}\sigma_{yk} } e^{\frac{y_{k,t}^2}{2 \sigma_{yk}^2}}  },
\end{equation}
\begin{equation}
\small \frac{P(\mathbf{X},\eta^*)}{P(\mathbf{X},\eta)} = \frac{ \prod_{t} \frac{1}{\sqrt{2\pi}\sigma_{yk} } e^{\frac{y_{k,t}^2}{2 \sigma_{yk}^2}} P(\mathbf{W}^1|K^1+1) P(K^1+1) }{ P(\mathbf{W}^1|K^1) P(K^1) },
\end{equation}
where $\frac{P(\mathbf{W}^1|K^1+1) }{P(\mathbf{W}^1|K^1) }$ is just the probability of generating a new column of $\mathbf{Z}^1$ with all zero values, specified by (2). And $\frac{P(K^1+1) }{P(K^1) }$ can be computed from Poisson distributions as the priors of IBP. As a result, we have
\begin{equation}
\small A(\eta^*,\eta) = min\left[1, \frac{ \frac{1}{K^1+1}   P(\mathbf{W}^1|K^1+1) P(K^1+1) }{  \frac{K^1_+}{K^1}  P(\mathbf{W}^1|K^1) P(K^1) }\right].
\end{equation}

Similarly, the proposal of delete a hidden factor with no linked edges is accepted with the probability
\begin{equation}
\small A(\eta^*,\eta) = min\left[1, \frac{ \frac{1}{K^1+1}   P(\mathbf{W}^1|K^1-1) P(K^1-1) }{  \frac{K^1_+}{K^1}  P(\mathbf{W}^1|K^1) P(K^1) }\right].
\end{equation}

To accomplish the algorithm, we need to sample $\mathbf{W}^1$ and $\mathbf{Y}^1$. Using the Gibbs sampling, we individually infer each variable of the two matrices in turn from the distributions $P(\mathbf{W}^1_{n,k}|\mathbf{X},\mathbf{W}^1_{-n,k},\mathbf{Y}^1)$ and $P(\mathbf{Y}^1_{k,t}|\mathbf{X},\mathbf{Y}^1_{-k,t},\mathbf{W}^1)$, where $\mathbf{W}^1_{-n,k}$ means all values of $\mathbf{W}^1$ except for $\mathbf{W}^1_{n,k}$ and $\mathbf{Y}^1_{-k,t}$ means all values of $\mathbf{Y}^1$ except for $\mathbf{Y}^1_{k,t}$. From the construction of our generative model and the Bayes' rule, we have
\begin{equation}
\begin{split}
\small P(\mathbf{W}^1_{n,k} & |\mathbf{X},\mathbf{W}^1_{-n,k},\mathbf{Y}^1) \\
 &\propto P(\mathbf{X}|\mathbf{W}^1_{n,k},\mathbf{W}^1_{-n,k},\mathbf{Y}^1) \cdot P(\mathbf{W}^1_{n,k}|\mathbf{W}^1_{-n,k}),
\end{split}
\end{equation}
where $P(\mathbf{X}|\mathbf{W}^1_{n,k},\mathbf{W}^1_{-n,k},\mathbf{Y}^1)$ is specified by the Gaussian likelihood we choose, and the term $P(\mathbf{W}^1_{n,k}|\mathbf{W}^1_{-n,k})$ can be obtained by integrating out the associated priors:
\begin{equation}
\begin{split}
& P(\mathbf{W}^1_{n,k}|\mathbf{W}^1_{-n,k}) = P(\mathbf{W}^1_{n,k}|\vec{\mathbf{W}^1}_{-n,k}) \\
& = \int_{\vec{\theta} \in S} P(\mathbf{W}^1_{n,k}|\vec{\theta}) P(\vec{\theta}|\vec{\mathbf{W}^1}_{-n,k}) \mathrm{d} \vec{\theta}, \\
& \vec{\theta}=(\sigma^2_{k},p_k); S=\{\sigma^2_{k} \in \Re^{+}, p_k \in [0,1] \},
\end{split}
\end{equation}
where $P(\mathbf{W}^1_{n,k}|\vec{\theta})$ is specified by (2) and $\vec{\mathbf{W}^1}_{-n,k}$ denotes all values of the $k$-th column of matrix $\mathbf{W}^1$ except for $\mathbf{W}^1_{n,k}$. Since the columns of $\mathbf{W}^1$ are generated independently, We compute $P(\mathbf{W}^1_{n,k}|\vec{\mathbf{W}^1}_{-n,k})$, instead of $P(\mathbf{W}^1_{n,k}|\mathbf{W}^1_{-n,k})$. Utilizing the Bayes' rule again, $P(\vec{\theta}|\mathbf{W}^1_{-n,k})$ can be computed by:
\begin{equation}
\begin{split}
P(\vec{\theta}|\vec{\mathbf{W}^1}_{-n,k}) \propto P(\vec{\mathbf{W}^1}_{-n,k}|\sigma^2_{k},p_k) \cdot P(\sigma^2_{k},p_k).
\end{split}
\end{equation}
The distribution $P(\vec{\mathbf{W}^1}_{-n,k}|\sigma^2_{k},p_k)$ can be computed by evaluating each element of the $k$-th column of matrix $\mathbf{W}^1$ except for $\mathbf{W}^1_{n,k}$ based on (2). The distribution $P(\sigma^2_{k},p_k)$ can be computed by multiplication of $P(\sigma^2_{k})$ and $P(p_k)$ which are specified by their priors defined in the generative model.

Similarly, we obtain the expression for $P(\mathbf{Y}^1_{k,t}|X,\mathbf{Y}^1_{-k,t},\mathbf{W}^1)$:
\begin{equation}
\begin{split}
& P(\mathbf{Y}^1_{k,t}|X,\mathbf{Y}^1_{-k,t},\mathbf{W}^1) \\
 &\propto P(X|\mathbf{Y}^1_{k,t},\mathbf{Y}^1_{-k,t},\mathbf{W}^1) \cdot P(\mathbf{Y}^1_{k,t}|\mathbf{Y}^1_{-k,t}),
\end{split}
\end{equation}
where $P(X|\mathbf{Y}^1_{k,t},\mathbf{Y}^1_{-k,t},\mathbf{W}^1)$ is specified by the Gaussian likelihood we choose. Since each element of $\mathbf{Y}^1_{.,t}$ is generated independently, $P(\mathbf{Y}^1_{k,t}|\mathbf{Y}^1_{-k,t})$ can be computed by its priors as (4).

The inference at the rest hidden layers is similar to the procedure used to infer the first hidden layer. Therefore, we summarize our inference algorithm in Algorithm 1.

\begin{algorithm}
\label{algoBayesianHidden}
\caption{MH steps for inferring first-layer hidden factors}
\For{$r=1,2, \ldots, \mbox{number of iterations} $}
{
  \For{$i=1,2, \ldots, N$ }
  {
    iteratively select column $k$ of $W$
    \eIf{$m_{-i,k}>0$}
    {
    propose adding a new hidden factor with probability specified by (12)
    }{
    propose deleting this hidden factor with probability specified by (13)
    }
    \For{$k=1,2, \ldots, K$  }
    {
    sample $W_{i,k}$ according to (14)
    }
    \For{ each element of $Y$  }
    {
    sample $Y_{k,t}$ according to (17)
    }
  }
}
\end{algorithm}

\section{Simulation Results and Discussions for Wireless Applications}
\label{sec:SimulationBayesianHidden}
We analyzed the performance of the proposed modified Metropolis-Hastings algorithm for inferring the true number of hidden factors in the first hidden layer. First, we fix the dimension of the observed data points, $N=16$, and vary the number of hidden factors, $K$, from 3 to 10. For each integer value of $K$, we generate a dataset containing $T=200$ data instances using the proposed generative model. Within one instance, $\mathbf{Y}^1$ is sampled according to its Gaussian prior. Then the weight matrix $\mathbf{W}^1$ is drawn from its distribution specified by (2). Finally, the data point $\mathbf{X}$ is generated by the Gaussian distribution where the parameters are expressed in terms of $\mathbf{Y}^1$ and $\mathbf{W}^1$. The rest model parameters are fixed at $\alpha'_{1}=3$ for the Beta distribution; $\alpha_{1} = 2$ and $\beta_{1} = 1$ for the Inversegamma distribution. The modified Metropolis-Hastings algorithm is initialized with three choices of $K$: $K=2$, $K=10$ or random positive integer between 3 and 10, and then runs for 200 iterations. Each dataset is estimated 10 times by the inference procedure described previously. We record the expectation of the estimated number of hidden factors and its variance as the result.

We plotted the results in Fig. 2. The modified Metropolis-Hastings algorithm is under the influence of initialization. When initializing $K=10$, which is much greater than the dimensions of the underlying model, the inferred $K$ values are generally much larger than the true values. However, when initializing $K$ randomly, the results correspondingly show some randomness. Another observation is that the MH method tends to over-estimate the number of hidden factors. This is because the proposal to add one hidden factor is preferred to be accepted. According to (10), the nominator is usually larger than the denominator because the denominator is composed by the multiplication of probability terms. Hence, the adding proposal is more likely to be accepted.

The proposed model can be utilized in unsupervised and nonparametric clustering problem in wireless networks. The estimated number of hidden factors solves one key challenge of clustering problem that is the determination of the number of clusters. In wireless security setting, the proposed model is a suitable solution to identify the attack devices in the communication system \cite{NamOnIdentify}. In the field of data analysis in the wireless networks, the proposed model can serve as the feature extraction approach \cite{deepwireless}. Moreover, the proposed model can contribute to the location estimation task in wireless networks \cite{madigan06location}. Many other wireless networking applications can be explored using the proposed framework.

\begin{figure}
\label{fig:finalresultBayesianHidden}
\centering
\includegraphics[width=85mm]{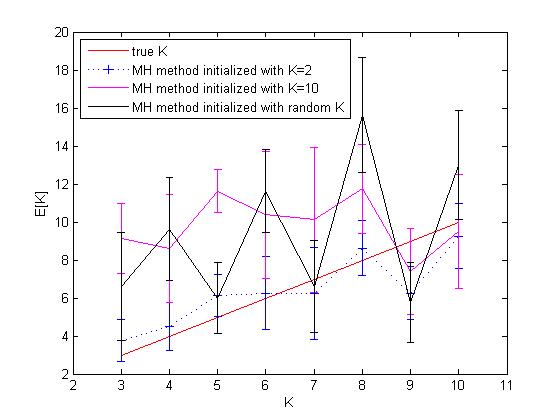}
\caption{Inferring the number of first-layer hidden factors using Metropolis-Hastings algorithm. Each curve shows the mean and variance of the expected value of the dimensionality $K$.}
\label{fig:infite model}
\end{figure}
\section{Conclusions}
\label{sec:ConclusionsBayesianHidden}

In this paper, we developed a deep hierarchical nonparametric Bayesian model to represent the underlying structure of observed data. Correspondingly, we proposed a modified Metropolis-Hastings algorithm to recover the number of hidden factors. Our simulation results on the hidden layer show that the algorithm discovers the model structure with some estimation errors. However, as shown in the results, our approach is capable of inferring increasing dimensions of hidden structures, which is due to the advantage of the nonparametric Bayesian technique. This indicates that the nonparametric Bayesian approach can be a suitable method for discovering complex structures.

\bibliographystyle{IEEETran}
\bibliography{./tempalte}

\begin{thebibliography}{10}
\providecommand{\url}[1]{#1}
\csname url@samestyle\endcsname
\providecommand{\newblock}{\relax}
\providecommand{\bibinfo}[2]{#2}
\providecommand{\BIBentrySTDinterwordspacing}{\spaceskip=0pt\relax}
\providecommand{\BIBentryALTinterwordstretchfactor}{4}
\providecommand{\BIBentryALTinterwordspacing}{\spaceskip=\fontdimen2\font plus
\BIBentryALTinterwordstretchfactor\fontdimen3\font minus
  \fontdimen4\font\relax}
\providecommand{\BIBforeignlanguage}[2]{{%
\expandafter\ifx\csname l@#1\endcsname\relax
\typeout{** WARNING: IEEEtran.bst: No hyphenation pattern has been}%
\typeout{** loaded for the language `#1'. Using the pattern for}%
\typeout{** the default language instead.}%
\else
\language=\csname l@#1\endcsname
\fi
#2}}
\providecommand{\BIBdecl}{\relax}
\BIBdecl

\bibitem{TehJor2010a}
Y.~W. Teh and M.~I. Jordan, ``Hierarchical {B}ayesian nonparametric models with
  applications,'' in \emph{Bayesian Nonparametrics: Principles and Practice},
  N.~Hjort, C.~Holmes, P.~M{\"u}ller, and S.~Walker, Eds.\hskip 1em plus 0.5em
  minus 0.4em\relax Cambridge University Press, 2010.

\bibitem{Thibaux07hierarchicalbeta}
R.~Thibaux and M.~I. Jordan, ``Hierarchical beta processes and the indian
  buffet process. this volume,'' In Practical Nonparametric and Semiparametric
  Bayesian Statistics, Tech. Rep., 2007.

\bibitem{Rasmussen00theinfinite}
C.~E. Rasmussen, ``The infinite gaussian mixture model,'' in \emph{In Advances
  in Neural Information Processing Systems 12}.\hskip 1em plus 0.5em minus
  0.4em\relax MIT Press, 2000, pp. 554--560.

\bibitem{Wood06anon-parametric}
F.~Wood, ``A non-parametric bayesian method for inferring hidden causes,'' in
  \emph{Proceedings of the Twenty-Second Conference on Uncertainty in
  Artificial Intelligence (UAI}.\hskip 1em plus 0.5em minus 0.4em\relax AUAI
  Press, 2006, pp. 536--543.

\bibitem{AdamsGraphical10}
H.~M.~W. Ryan P.~Adams and Z.~Ghahramani, ``Learning the structure of deep
  sparse graphical models,'' in \emph{13-th International Conference on
  Artificial Intelligence and Statistics}, Chia Laguna, Sardinia, Italy, May
  2010.

\bibitem{ICML2011Chen251}
B.~Chen, G.~Polatkan, G.~Sapiro, L.~Carin, and D.~B. Dunson, ``The hierarchical
  beta process for convolutional factor analysis and deep learning,'' in
  \emph{Proceedings of the 28th International Conference on Machine Learning
  (ICML-11)}, L.~Getoor and T.~Scheffer, Eds.\hskip 1em plus 0.5em minus
  0.4em\relax New York, NY, USA: ACM, 2011, pp. 361--368.

\bibitem{rai08ihfrm}
P.~Rai and H.~{Daum\'e III}, ``The infinite hierarchical factor regression
  model,'' in \emph{Proceedings of the Conference on Neural Information
  Processing Systems (NIPS)}, Vancouver, Canada, 2008.

\bibitem{Knowles07}
D.~Knowles and Z.~Ghahramani, ``Infinite sparse factor analysis and infinite
  independent components analysis,'' in \emph{Independent Component Analysis
  and Signal Separation}, ser. Lecture Notes in Computer Science, M.~Davies,
  C.~James, S.~Abdallah, and M.~Plumbley, Eds.\hskip 1em plus 0.5em minus
  0.4em\relax Springer Berlin Heidelberg, 2007, vol. 4666, pp. 381--388.

\bibitem{Lee09convDBNAudio}
H.~Lee, P.~T. Pham, Y.~Largman, and A.~Y. Ng, ``Unsupervised feature learning
  for audio classification using convolutional deep belief networks,'' in
  \emph{Advances in Neural Information Processing Systems 22: 23rd Annual
  Conference on Neural Information Processing Systems}, Vancouver, Canada,
  December 2009.

\bibitem{NamCellularBook}
N.~T. Nguyen, X.~Liu, and R.~Zheng, ``A nonparametric bayesian approach for
  opportunistic data transfer in cellular networks,'' in \emph{in Proceedings
  of the 7th International Conference of Wireless Algorithms, Systems, and
  Applications (WASA)}, Yellow Mountain, China, August 2012, pp. 88--99.

\bibitem{Griffiths11IBP}
T.~L. Griffiths and Z.~Ghahramani, ``The indian buffet process: An introduction
  and review,'' \emph{J. Mach. Learn. Res.}, vol.~12, pp. 1185--1224, July
  2011.

\bibitem{FaceBayesianCVPR2012}
L.~Ma, C.~Wang, B.~Xiao, and W.~Zhou, ``Sparse representation for face
  recognition based on discriminative low-rank dictionary learning,'' in
  \emph{Computer Vision and Pattern Recognition (CVPR), 2012 IEEE Conference
  on}, Providence, RI, June 2012, pp. 2586--2593.

\bibitem{Hinton06FLA}
G.~E. Hinton, S.~Osindero, and Y.-W. Teh, ``A fast learning algorithm for deep
  belief nets,'' \emph{Neural Comput.}, vol.~18, no.~7, pp. 1527--1554, July
  2006.

\bibitem{carreiraperpin:aistats2005}
M.~A. Carreira-Perpi{\~n} and G.~Hinton, ``On contrastive divergence
  learning,'' R.~G. Cowell and Z.~Ghahramani, Eds.\hskip 1em plus 0.5em minus
  0.4em\relax Society for Artificial Intelligence and Statistics, 2005, pp.
  33--40, (Available electronically at
  \mbox{http://www.gatsby.ucl.ac.uk/aistats/}).

\bibitem{NamOnIdentify}
N.~Nguyen, R.~Zheng, and Z.~Han, ``On identifying primary user emulation
  attacks in cognitive radio systems using nonparametric bayesian
  classification,'' \emph{Signal Processing, IEEE Transactions on}, vol.~60,
  no.~3, pp. 1432--1445, March 2012.

\bibitem{deepwireless}
S.~Chinchali and S.~Tandon, ``Location estimation in wireless networks: A
  bayesian approach,'' (Online report at
  http://cs229.stanford.edu/projects2012.html).

\bibitem{madigan06location}
D.~Madigan, W.-H. Ju, P.~Krishnan, A.~S. Krishnakumar, and I.~Zorych,
  ``Location estimation in wireless networks: a bayesian approach,''
  \emph{Statistica Sinica}, vol.~16, no.~2, pp. 495--522, 2006.

\end{thebibliography}
\end{document}